\documentclass{article}

\usepackage[preprint]{corl_2026} % Preprint version for arXiv.
\usepackage[utf8]{inputenc}
\usepackage[T1]{fontenc}
\usepackage{authblk}
\usepackage{amsmath, amssymb, amsfonts}
\usepackage{graphicx}
\usepackage{booktabs}
\usepackage{placeins}
\usepackage{hyperref}
\usepackage{geometry}
\usepackage{cite}
\usepackage{algorithm}       
\usepackage{algpseudocode}   
\usepackage{booktabs} 
\usepackage{array}

\title{Diffusion Policies for Short-Horizon Planning in Robot Crowd Navigation}

\author{
  Wendong Li \qquad Jochen Garcke\\
  Institute for Numerical Simulation\\
  University of Bonn\\
  Bonn, Germany
}

\begin{document}
\maketitle

%===============================================================================

\begin{abstract}
Robot crowd navigation requires safe and efficient decision-making under dense, dynamic, and multimodal human--robot interactions. Existing reinforcement-learning methods typically output a single reactive action at each timestep, which limits their ability to represent diverse short-term avoidance strategies. We propose Planning Diffusion Policy Optimization (PDPO), an offline-to-online reinforcement-learning framework that uses a diffusion policy to generate short-horizon action chunks for crowd navigation. PDPO is first pretrained on collision-avoidance demonstrations and then fine-tuned online with PPO by treating the denoising process as an internal decision process. During execution, the policy generates a five-step action chunk and applies it in a receding-horizon manner. Furthermore, we observe an evaluation artifact in common crowd-navigation benchmarks: without explicit boundary constraints, learned agents may leave the valid domain and bypass dense crowds. To address this, we introduce a setting in which boundary violations are treated as collisions. Experiments show that PDPO obtains an improved success rate over strong baselines, and ablations demonstrate that action chunks are especially important for the modified bounded benchmark.
\end{abstract}

%===============================================================================

\section{Introduction}
\label{sec:introduction}

Robot crowd navigation is a fundamental problem in robotics, where a robot must reach its goal safely and efficiently while moving among dense, dynamic, and partially observable crowds~\citep{DBLP:conf/iros/EverettCH18,DBLP:conf/icra/ChenLKA19}. 
The difficulty of this problem lies not only in collision avoidance, but also in the multi-modal nature of human--robot interactions. 
When approaching pedestrians, multiple maneuvers may be simultaneously feasible, such as passing on different sides, slowing down to yield, or moving through a temporary gap. 
Effective navigation therefore requires both representing diverse local decisions and maintaining temporal consistency over the next few steps.

Classical rule-based methods, such as ORCA and Social Force, compute collision-avoidance actions from hand-designed local interaction rules~\citep{van2011reciprocal,Helbing1995SocialFM}. 
Although these methods are efficient and widely used, they are often reactive and can become overly conservative in dense crowds, leading to freezing behaviors~\citep{DBLP:conf/iros/TrautmanK10}. 
Learning-based methods, especially deep reinforcement learning, improve adaptability by optimizing long-term returns from interaction experience~\citep{DBLP:conf/icra/ChenLKA19,DBLP:conf/icra/LiuCLCC21,DBLP:conf/icra/LiuCHCHLMGD23,DBLP:conf/icra/ZhouG24,DBLP:conf/iros/ChenHNMS20}. 
However, many reinforcement-learning-based navigation policies implicitly adopt a single-step action formulation, where the policy predicts one action per timestep and is often parameterized by a simple unimodal distribution such as a Gaussian.
Such a representation is restrictive for crowd navigation, where different avoidance strategies can be valid under the same observation. 
Moreover, while future outcomes are considered implicitly through accumulated rewards or value estimation, the policy itself usually does not explicitly represent a short-horizon action plan.

To address these limitations, we introduce \textit{Planning Diffusion Policy Optimization} (PDPO), an offline-to-online reinforcement learning framework that brings diffusion-policy-based reinforcement learning to crowd navigation.
Although diffusion policies and action chunking have been studied in other control settings \citep{DBLP:conf/iclr/RenLAS0MBDS25,DBLP:journals/ijrr/ChiXFCDBTS25}, dense crowd navigation poses distinct challenges beyond structured manipulation-style control, due to dynamic multi-agent interactions and safety-critical collision avoidance.
PDPO addresses these challenges by generating a short-horizon action chunk conditioned on the current crowd observation, which serves as a local motion plan rather than only a sequence of low-level motor commands.
Following a receding-horizon scheme, only the first action is executed before replanning at the next timestep.
This design enables the policy to capture multi-modal navigation choices while promoting temporally coherent motion in dynamic crowds.

Following the common two-stage training paradigm of diffusion policies~\citep{DBLP:conf/iclr/RenLAS0MBDS25,DBLP:journals/ijrr/ChiXFCDBTS25,DBLP:conf/nips/HoJA20}, PDPO first pretrains the diffusion policy with behavioral cloning on demonstration trajectories with effective collision-avoidance behaviors, providing a reasonable initialization for navigation.
It then fine-tunes the policy online with reinforcement learning to improve task performance beyond the demonstrations and adapt to the interaction distribution induced by its own decisions.
Together, PDPO provides a practical and effective diffusion-based planning framework for dense crowd navigation.

In addition to policy design, we find that the evaluation protocol itself can substantially affect crowd-navigation results. 
In commonly used simulation benchmarks~\citep{DBLP:conf/icra/LiuCHCHLMGD23}, the lack of explicit boundary constraints can allow a robot to leave the valid navigation region and bypass dense crowds, creating unintended shortcut behaviors. 
Such artifacts may lead to overly optimistic evaluations and obscure the true crowd-awareness of learned policies. 
To obtain a more faithful evaluation, we introduce a bounded benchmark setting in which leaving the workspace is treated as a collision, and we compare policies under both the original and corrected environments.

The main contributions of this work are threefold:

\textbf{Diffusion policies for crowd navigation.}
We introduce a diffusion-policy formulation for robot crowd navigation, enabling the policy to represent non-Gaussian and multimodal avoidance behaviors that are difficult to capture with commonly used single-step unimodal Gaussian policies.

\textbf{Action-chunk-based short-horizon planning.}
Instead of producing a single reactive action, PDPO generates a short sequence of velocity commands and executes the first action in a receding-horizon manner. This action-chunk representation provides a local motion plan while preserving responsiveness to dynamic crowds.

\textbf{Bounded domain as a modified benchmark.}
We identify an evaluation artifact in commonly used crowd-navigation experiments: without explicit workspace boundary constraints, RL agents may leave the nominal navigation region and thereby bypass dense interactions. We introduce a bounded CrowdNav benchmark and observe that in this modified setting the performance substantially differs.

The code and benchmark setup will be made available on publication.
%===============================================================================
\section{Related Work}

    \paragraph{Rule-based Crowd Navigation.}
    Traditional crowd navigation largely relies on rule-based methods. 
    Notable examples include ORCA~\citep{van2011reciprocal}, which enforces collision avoidance via geometric constraints, and the Social Force model~\citep{Helbing1995SocialFM}, which simulates pedestrian motion using attraction–repulsion forces. 
    These methods compute actions directly from observations based on predefined rules, without learning from data. 
    As a result, they are reactive and local, often performing poorly in dense crowds where conservative behavior may stall the robot.
    
   \paragraph{Learning-based Methods.}
    Early learning-based crowd navigation relied on deep value-based RL, estimating state or state--action values to guide decisions~\citep{DBLP:conf/icra/ChenLKA19,DBLP:conf/iros/ChenHNMS20,DBLP:conf/icra/ShiZZLB22,DBLP:conf/icra/ZhouG24}.
    While effective in many benchmark settings, value-based methods are less direct for continuous action control and complex human--robot interactions.
    Policy optimization approaches, such as PPO-based crowd-navigation policies, have therefore become widely used due to their compatibility with continuous actions and neural network function approximation~\citep{DBLP:conf/icra/LiuCLCC21,DBLP:conf/icra/LiuCHCHLMGD23,DBLP:journals/corr/SchulmanWDRK17}.
    However, these methods typically optimize policies at the single-action level, so temporally coherent short-horizon motion plans are not explicitly represented but are instead encouraged indirectly through value estimation, rollout-based returns, or reward design.
    
    More recent works such as CrowdNav++~\citep{DBLP:conf/icra/LiuCHCHLMGD23} incorporate human trajectory prediction into reward design to improve safety and social compliance.
    These methods improve navigation performance, but they typically retain a single-step policy formulation and often implicitly rely on unimodal Gaussian action parameterizations.
    As a result, they do not explicitly model a distribution over short-horizon action sequences, which is the focus of our work.
    Zhang et al.~\citep{yao2025towards} further improved safety under prediction errors through conformal uncertainty estimation and constrained policy optimization; their focus, however, remains on safe policy optimization rather than explicit short-horizon action-sequence modeling.
    
    Feature extraction architectures have also evolved: from MLP-based pooling in SARL~\citep{DBLP:conf/icra/ChenLKA19}, to spatio-temporal graphs with RNNs in DSRNN~\citep{DBLP:conf/icra/LiuCLCC21,DBLP:conf/icra/ZhouG24}, and further to self-attention with implicit temporal encoding in CrowdNav++~\citep{DBLP:conf/icra/LiuCHCHLMGD23}. 
    In contrast to these architecture-focused improvements, our work focuses on the policy representation itself, using diffusion-based action chunks to model multi-modal short-horizon navigation behaviors.

    \paragraph{Diffusion Policies.}
    Recent advances in diffusion models, particularly denoising diffusion probabilistic models (DDPMs)~\citep{DBLP:conf/nips/HoJA20}, have motivated their use as expressive policy classes for decision making and robotic control.
    In robotic imitation learning, diffusion policies have been used to model complex action distributions and generate multi-step action sequences through iterative denoising~\citep{DBLP:journals/ijrr/ChiXFCDBTS25}.
    Recent work further extends diffusion policies to reinforcement learning by treating the denoising process as an MDP and optimizing it with policy-gradient methods such as PPO~\citep{DBLP:conf/iclr/RenLAS0MBDS25,DBLP:journals/corr/SchulmanWDRK17}.
    These methods commonly follow a two-stage pipeline, where the policy is first pretrained from demonstrations and then refined with goal-conditioned or reward-based fine-tuning~\citep{DBLP:conf/iclr/RenLAS0MBDS25,DBLP:journals/ijrr/ChiXFCDBTS25}.
    Building on this paradigm, PDPO adapts diffusion-based short-horizon action generation to dense crowd navigation, enabling temporally coherent local planning under dynamic multi-agent interactions.
%===============================================================================

\section{Preliminaries}
\label{sec:preliminaries}

\subsection{Problem Formulation}
\label{sec:problem_formulation}

As in prior learning-based crowd-navigation work~\citep{DBLP:conf/icra/ChenLKA19,DBLP:conf/icra/LiuCHCHLMGD23}, crowd navigation is formulated as a Markov Decision Process (MDP)
$(\mathcal{S}, \mathcal{A}, \mathcal{P}_0, \mathcal{P}, \mathcal{R})$.
At timestep $t$, the robot observes its own state $w^t$, including its position, velocity, goal, maximum speed, orientation, and radius.
For each observable human $i$, the observation includes the current position $u_i^t$ and predicted future positions $\hat{u}_i^{t+1:t+L}$ over a short horizon $L$.
The state is represented as
\begin{equation}
    s_t = [w^t, h_1^t, \ldots, h_{n_t}^t],
    \quad
    h_i^t = [u_i^t, \hat{u}_i^{t+1:t+L}],
\end{equation}
where $n_t$ is the number of humans observed at timestep $t$.

The robot action is a two-dimensional velocity command $a_t \in \mathbb{R}^2$.
At each timestep, the robot receives reward $r_t = \mathcal{R}(s_t,a_t)$ and transitions to $s_{t+1} \sim \mathcal{P}(\cdot \mid s_t,a_t)$.
Humans follow a fixed policy that is unknown to the robot.
An episode terminates when the robot reaches its goal, collides with a human or boundary, or exceeds the maximum episode length.

Following CrowdNav++~\citep{DBLP:conf/icra/LiuCHCHLMGD23}, we largely adopt its reward design and use a constant-velocity predictor for short-horizon human position prediction.
Except for boundary handling in the bounded CrowdNav, these components are fixed across learning-based methods to isolate the effect of policy representation; details are provided in Appendix~\ref{app:architecture_hyperparameters}.

\subsection{Diffusion Policy Background}
\label{sec:diffusion_policy_background}

Most policy-gradient methods for continuous control parameterize the policy as a Gaussian distribution,
\begin{equation}
    \pi_\theta(a \mid s)
    =
    \mathcal{N}\big(a; \mu_\theta(s), \Sigma_\theta(s)\big),
    \label{eq:gaussian_policy}
\end{equation}
where $\mu_\theta(s)$ and $\Sigma_\theta(s)$ are predicted by a neural network.
This formulation provides an explicit likelihood and is convenient for PPO-style policy optimization~\citep{DBLP:journals/corr/SchulmanWDRK17}, 
but restricts the conditional action distribution to a unimodal family, which can be limiting in tasks where multiple actions or action sequences may be feasible under the same observation.

A diffusion policy instead defines the conditional distribution implicitly through an iterative denoising process~\citep{DBLP:conf/nips/HoJA20}.
Let $x_0$ denote the policy output, which may be either a single action or a sequence of actions.
Starting from Gaussian noise $x_K \sim \mathcal{N}(0,I)$, the policy applies $K$ reverse denoising transitions,
\begin{equation}
    p_\theta(x_{k-1} \mid x_k, s)
    =
    \mathcal{N}\!\left(
        x_{k-1};
        \mu_\theta(x_k, s, k),
        \sigma_k^2 I
    \right),
    \quad k=K,\ldots,1.
    \label{eq:diffusion_reverse}
\end{equation}
Equivalently, the mean $\mu_\theta$ can be parameterized through a noise-prediction network $\varepsilon_\theta(x_k,s,k)$.
Through this iterative denoising process, the induced policy $\pi_\theta(x_0 \mid s)$ can represent complex, non-Gaussian, and multi-modal action distributions~\citep{DBLP:journals/ijrr/ChiXFCDBTS25}.
In PDPO, we instantiate $x_0$ as a short action chunk and optimize the diffusion policy with behavioral cloning following online RL~\citep{DBLP:conf/iclr/RenLAS0MBDS25}.

%===============================================================================

\section{Planning Diffusion Policy Optimization}
\label{sec:method}

PDPO is an offline-to-online reinforcement learning framework that uses a diffusion policy to generate short-horizon action chunks for crowd navigation.
At each timestep, the robot encodes the current crowd observation $s_t$ into a latent representation $z_t$, samples a five-step action chunk conditioned on $z_t$, executes the first action, and replans at the next timestep.
The diffusion policy is first pretrained from collision-avoidance demonstrations and then fine-tuned online with task rewards.

\paragraph{Crowd Observation Encoding.}
\label{sec:observation_encoding}

Our crowd observation encoder is adapted from prior crowd-navigation architectures~\citep{DBLP:conf/icra/LiuCHCHLMGD23,DBLP:conf/icra/ZhouG24}.
It maps the crowd observation $s_t$ to a latent representation $z_t$ that conditions the diffusion policy.
The encoder uses MLP layers for human-feature embedding, masked self-attention over observable humans, and attention pooling to aggregate human--robot interaction features.
A visibility mask excludes unobservable humans from the attention computation.
Architecture details and hyperparameters are provided in Appendix~\ref{app:architecture_hyperparameters}.

\paragraph{Diffusion Action Chunk Policy.}
\label{sec:diffusion_action_chunk}

Given the latent observation $z_t$, PDPO samples a five-step action chunk
\begin{equation}
x_0^t = \left[x_{0,1}^t, x_{0,2}^t, x_{0,3}^t, x_{0,4}^t, x_{0,5}^t\right],
\end{equation}
where each $x_{0,j}^t \in \mathbb{R}^2$ is a velocity command.
The chunk is generated by the diffusion policy introduced in Sec.~\ref{sec:diffusion_policy_background}, conditioned on $z_t$.

The action chunk provides an explicit short-horizon representation of local motion.
In crowd navigation, meaningful avoidance behaviors are often defined over several consecutive actions rather than a single velocity command, such as passing a pedestrian on one side, slowing down to yield, or moving through a temporary gap.
Modeling a distribution over action chunks allows the policy to capture multi-modal motion patterns at the sequence level while encouraging consistency among consecutive actions within the same local plan.

During execution, only the first action is applied to the environment:
\begin{equation}
a_t = x_{0,1}^t.
\end{equation}
At the next timestep, the robot receives a new observation and samples a new action chunk.
This receding-horizon execution combines short-horizon planning with closed-loop feedback: the chunk provides a temporally coherent local motion pattern, while replanning keeps the robot responsive to changes in the crowd.

\paragraph{Offline Demonstration Pretraining.}
\label{sec:pretraining}

Training a diffusion policy directly from online reinforcement learning is challenging in crowd navigation, since the policy must learn a multi-step denoising process over action chunks while exploring safely in dense dynamic environments.
We therefore first pretrain the diffusion policy with behavioral cloning on collision-avoidance demonstrations, following prior diffusion-policy methods that initialize action-generation models from demonstration data before downstream deployment or optimization~\citep{DBLP:conf/iclr/RenLAS0MBDS25,DBLP:journals/ijrr/ChiXFCDBTS25}.

Demonstration trajectories are collected in the crowd simulation environment by rolling out ORCA.
At each timestep, we record the observation and executed velocity command, and construct an expert action chunk from the velocity commands at the current timestep and the following $H-1$ timesteps.
We use $H=5$ in our experiments.

The diffusion policy is pretrained with the standard denoising objective.
Given an expert chunk $x_0^t$, we sample a diffusion step $k$ and Gaussian noise $\epsilon$, perturb the chunk into $x_k^t$, and train the denoising network to predict the injected noise:
\begin{equation}
    \mathcal{L}_{\mathrm{BC}}(\theta)
    =
    \mathbb{E}_{t,k,\epsilon}
    \left[
    \left\|
    \epsilon - \epsilon_\theta(x_k^t, z_t, k)
    \right\|_2^2
    \right].
    \label{eq:bc_loss}
\end{equation}
This stage provides a reasonable initialization with feasible collision-avoidance behaviors, which is then refined through online reinforcement learning.

\paragraph{Online Reinforcement Learning Fine-tuning.}
\label{sec:online_finetuning}

After pretraining, we fine-tune the diffusion policy online with PPO under task rewards, following recent diffusion-policy optimization approaches~\citep{DBLP:conf/iclr/RenLAS0MBDS25}.
The reverse denoising process is treated as an internal MDP: during rollout collection, we store denoising transitions and their log probabilities for the generated chunk.
After the first action of the chunk is executed in the environment, the resulting reward and advantage are assigned to the denoising transitions that produced the chunk.

The policy is then updated with the PPO objective at the denoising-step level.
The full augmented-MDP formulation, advantage assignment, and implementation details are provided in Appendix~\ref{app:diffusion_ppo}.
\section{Experiment}
\label{sec:experiment}

\subsection{Simulation Environment}

All experiments are conducted in the CrowdNav++ simulator~\citep{DBLP:conf/icra/LiuCHCHLMGD23}, 
a recent and challenging benchmark for robot crowd navigation that evaluates policies in dense, dynamic human crowds.
The environment is a $12 \times 12~\mathrm{m}^2$ two-dimensional workspace containing one robot and 20 human agents.
The task requires the robot to reach its goal while avoiding collisions and maintaining socially acceptable navigation behavior in dense crowds.

At the beginning of each episode, the robot and all humans are initialized with random positions and goals.
Human agents follow ORCA for collision avoidance among humans, but they do not observe or react to the robot.
The ORCA policy used by humans is not accessible to the robot during training or execution.
The robot is modeled as a circular agent with radius $0.2~\mathrm{m}$, maximum speed $1~\mathrm{m/s}$, and sensing range $6~\mathrm{m}$.
Human radii are uniformly sampled from $[0.3, 0.5]~\mathrm{m}$, and their maximum speeds are sampled from $[0.5, 1.5]~\mathrm{m/s}$.

To generate dynamic crowd flows, human agents are assigned new goal locations upon reaching their current goals.
Additionally, at each second, each human independently changes its goal with probability $0.5$.
Since humans do not react to the robot, the robot cannot exploit cooperative human responses through overly aggressive navigation.

Each episode terminates when the robot reaches its goal, collides with a human, or exceeds the maximum episode length of 50 seconds.
Episodes that reach the time limit are treated as timeouts.

\paragraph{Bounded CrowdNav setting.}
We observe that the original CrowdNav++ benchmark~\citep{DBLP:conf/icra/LiuCHCHLMGD23} does not enforce explicit boundary constraints, allowing the robot to leave the workspace region and thereby bypass dense crowds.
This artifact may lead to overly optimistic evaluation of learned policies.
To address this, we introduce boundary constraints by adding walls around the environment.
While using the same crowd-generation process, robot dynamics, and evaluation metrics,  boundary violations are now treated as collisions. 
% Any attempt to leave the workspace is treated as a collision.
We refer to this modified environment as \textbf{Bounded CrowdNav}.

% \subsection{Evaluation Settings}

% We evaluate methods under two settings.
% The \textbf{original benchmark} follows the standard CrowdNav++ protocol without explicit boundary constraints, allowing comparison with previously reported results.
% \textbf{Bounded CrowdNav} uses the same crowd-generation process, robot dynamics, and evaluation metrics, but treats workspace boundary violations as collisions.
% We use Bounded CrowdNav as the main controlled evaluation setting because it removes shortcut behaviors caused by leaving the valid workspace.

\subsection{Experimental Setup and Training}

\paragraph{Baselines and ablations.}
We compare \textbf{Planning Diffusion Policy Optimization (PDPO)} with two representative crowd-navigation baselines:
\begin{itemize}
    \item \textbf{ORCA}: a rule-based collision-avoidance planner, also used to generate demonstration data for imitation pretraining~\citep{van2011reciprocal}.
    \item \textbf{CrowdNav++ (Const. Vel.)}: a strong learning-based crowd-navigation baseline using a constant-velocity human trajectory predictor~\citep{DBLP:conf/icra/LiuCHCHLMGD23}.
\end{itemize}

To isolate the effect of action chunking, we include one ablation:
\begin{itemize}
    \item \textbf{PDPO w/o action chunk}: a diffusion-policy variant that generates a single action per timestep instead of a multi-step action chunk.
\end{itemize}

Unless noted otherwise, all learning-based methods use the same constant-velocity predictor to forecast human motion over the next five timesteps.

\paragraph{Training details.}
PDPO and its ablation variants are trained with the same two-stage offline-to-online procedure.
First, the diffusion policy is pretrained via behavioral cloning using approximately $3{,}000$ demonstration episodes collected in simulation.
The pretraining stage runs for $200$ optimization iterations.
The policy is then fine-tuned online for around $12$ million environment timesteps.
We use the same training budget for PDPO and its ablations.
PDPO uses chunk length $H=5$, corresponding to a $1.25\,\mathrm{s}$ planning horizon under the $0.25\,\mathrm{s}$ control interval.

Crowd-navigation benchmarks are sensitive to random seeds.
Following CrowdNav++~\citep{DBLP:conf/icra/LiuCHCHLMGD23}, we use separate random seed sets for training, validation, and testing.
Each final policy is evaluated on $500$ unseen test seeds, and results are averaged over these episodes.

\paragraph{Training under two different benchmark settings.}
% We evaluate methods under two settings.
The \textbf{original benchmark} follows the standard CrowdNav++ protocol without explicit boundary constraints, allowing comparison with previously reported results.
We use \textbf{Bounded CrowdNav} as the main controlled evaluation setting because it removes shortcut behaviors caused by leaving the valid workspace.
For the original benchmark, we report ORCA and CrowdNav++ results from the CrowdNav++ paper~\citep{DBLP:conf/icra/LiuCHCHLMGD23}, since the environment and evaluation protocol are unchanged.
For Bounded CrowdNav, all learning-based methods, including PDPO, its ablation, and CrowdNav++, are trained and evaluated under the same bounded environment.

\paragraph{Evaluation metrics.}
We evaluate navigation performance using standard crowd-navigation metrics that measure efficiency, safety, and social compliance: success rate (SR, \%), navigation time (NT, s), path length (PL, m), intrusion rate (ITR, \%), and social distance (SD, m).
ITR measures how often the robot enters human personal-space regions, while SD measures the average closest social distance during navigation.

\subsection{Results on the Original Benchmark}

\begin{table}[htbp]
\centering
\caption{Navigation performance on the original CrowdNav++ benchmark without explicit boundary constraints. ORCA and CrowdNav++ results are reported from~\citep{DBLP:conf/icra/LiuCHCHLMGD23}.}
\label{tab:original_results}
\begin{tabular}{lccccc}
\toprule
\textbf{Method} & \textbf{SR}$\uparrow$ & \textbf{NT}$\downarrow$ & \textbf{PL}$\downarrow$ & \textbf{ITR}$\downarrow$ & \textbf{SD}$\uparrow$ \\ 
\midrule
ORCA & 69.0 & 14.77 & 17.67 & 19.61 & 0.38 \\
CrowdNav++ (Const. Vel.) & 87.0 & 14.03 & 20.14 & 7.00 & 0.42 \\
\midrule
PDPO w/o action chunk & 86.2 & 14.59 & 19.20 & 8.04 & 0.40 \\
PDPO (Ours) & 90.6 & 14.50 & 20.25 & 7.22 & 0.41 \\
\bottomrule
\end{tabular}
\end{table}

Table~\ref{tab:original_results} reports performance on the original CrowdNav++ benchmark.
PDPO achieves a success rate of $90.6\%$, outperforming the reported CrowdNav++ result by $3.6$ percentage points.
The single-action diffusion-policy variant performs comparably to CrowdNav++, while action chunking improves the success rate from $86.2\%$ to $90.6\%$.
These results suggest that diffusion-based action chunks can improve navigation performance even under the original benchmark protocol.

To quantify the effect of missing boundary constraints, we further analyze successful trajectories under the original benchmark and measure the fraction in which the robot leaves the nominal workspace.
Using the same evaluation scenes, we find that $16.1\%$ of successful CrowdNav++ trajectories, $31.8\%$ of successful ORCA trajectories, $18.7\%$ of successful PDPO trajectories without action chunks, and $19.8\%$ of successful PDPO trajectories go out of bounds while still being counted as successes.
This shows that boundary handling can substantially affect evaluation outcomes, motivating Bounded CrowdNav as our main controlled setting.

\subsection{Results on Bounded CrowdNav}

\begin{table}[htbp]
\centering
\caption{Navigation performance on \emph{Bounded CrowdNav}. All learning-based methods are trained and evaluated under the same bounded environment, where learning-based methods are evaluated over three random seeds for training the model. ORCA is deterministic and reported without standard deviation. The standard deviation for SD is negligible for all approaches.}

\label{tab:bounded_results}
\begin{tabular}{lccccc}
\toprule
\textbf{Method} & \textbf{SR}$\uparrow$ & \textbf{NT}$\downarrow$ & \textbf{PL}$\downarrow$ & \textbf{ITR}$\downarrow$ & \textbf{SD}$\uparrow$ \\ 
\midrule
ORCA & 47.0 & 21.26 & 17.14 & 1.35 & 0.50 \\
CrowdNav++ (Const. Vel.) & 74.5 $\pm$ 2.1 & 13.35 $\pm$ 1.13 & 18.42 $\pm$ 0.65 & 8.53 $\pm$ 0.30 & 0.42\\% $\pm$ 0.01 \\
\midrule
PDPO w/o action chunk & $73.7 \pm 0.7$ & $13.05 \pm 0.46$ & $20.93 \pm 0.50$ & $7.21 \pm 0.74$ & $0.40$ \\ % \pm 0.00$ \\
PDPO (Ours) & $84.7 \pm 1.4$ & $10.92 \pm 0.04$ & $20.58 \pm 0.19$ & $6.81 \pm 0.04$ & $0.41$ \\ % \pm 0.00$\\
\bottomrule
\end{tabular}
\end{table}

Table~\ref{tab:bounded_results} reports results on \emph{Bounded CrowdNav}.
PDPO achieves a success rate of $84.7\%$, outperforming the retrained CrowdNav++ by $11.7$ percentage points and the single-action diffusion-policy ablation by $11.0$ percentage points on average.
This indicates that PDPO's improvement is not merely due to the diffusion-policy parameterization, but largely comes from generating short-horizon action chunks.

The bounded setting also exposes the impact of boundary handling: CrowdNav++ drops from its reported $87.0\%$ success rate in the original benchmark to around $74.5\%$ under boundary constraints.
PDPO also achieves the lowest intrusion rate, reducing ITR from $8.53\%$ for CrowdNav++ to $6.81\%$.
Although its path length is longer than that of CrowdNav++, which is a reasonable trade-off in dense crowds, where safer trajectories may require detours, the navigation time is shorter.
Together, these results show that action-chunk diffusion policies improve both task success and crowd-aware safety.

\subsection{Visualization of Action Chunks}

We provide qualitative visualizations to illustrate how PDPO uses action chunks for short-horizon planning in dense crowds.
At each decision step, the policy generates a 5-step action chunk with a control interval of $0.25\,\mathrm{s}$, corresponding to a $1.25\,\mathrm{s}$ local plan.
Only the first action is executed before replanning.
\begin{figure}[t]
    \centering
    \begin{tabular}{ccc}
        \includegraphics[width=0.31\textwidth]{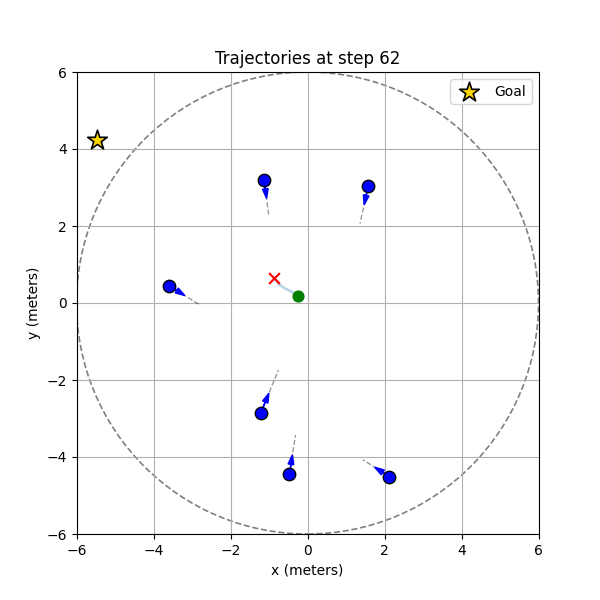} &
        \includegraphics[width=0.31\textwidth]{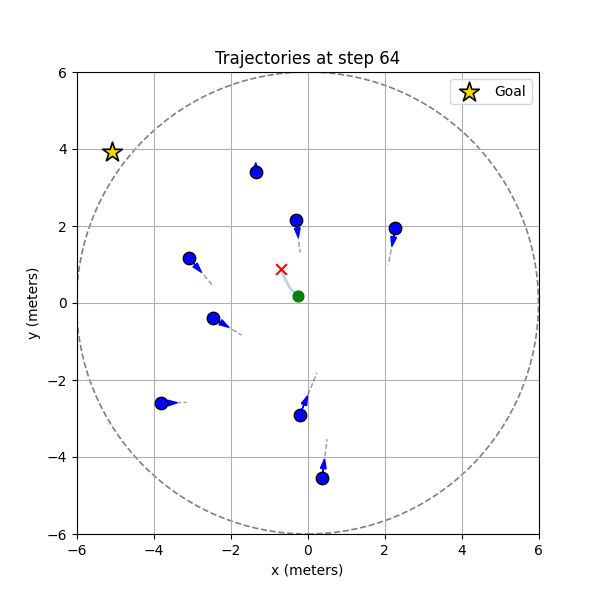} &
        \includegraphics[width=0.31\textwidth]{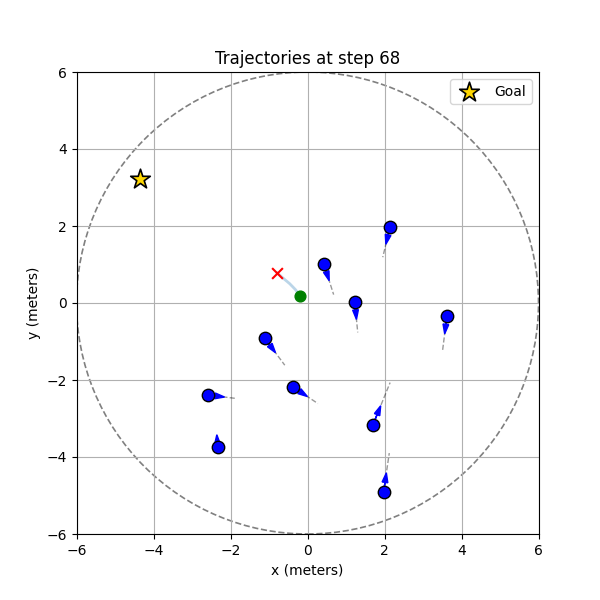} \\
        \small Step 62 & \small Step 64 & \small Step 68
    \end{tabular}
    \caption{Representative snapshots of action chunks generated by PDPO during navigation through a dense crowd. 
In each snapshot, the dashed circle denotes the robot's observation range, blue dots denote humans with predicted linear trajectories, the green dot denotes the robot, and the red cross the end of the action chunk.
    The generated chunks provide coherent short-horizon local plans that adapt to nearby pedestrians.
    Complete storyboards and multimodal samples are in Appendix~\ref{app:action_chunk_visualization}.}
    \label{fig:action_chunks_main}
\end{figure}
Figure~\ref{fig:action_chunks_main} shows representative snapshots from a dense navigation episode.
As nearby pedestrians enter the robot's observation range, the generated chunks adjust toward available gaps while maintaining short-horizon motion consistency.
Around step 64, the chunk exhibits a local turning behavior in response to surrounding pedestrians.
By step 68, after the robot moves away from the densest interaction region, the planned motion becomes longer again, indicating faster progress toward the goal.
%===============================================================================
\section{Limitations and Future Work}
\label{sec:limitations}

% This work has several limitations. 

PDPO uses an iterative denoising process to generate action chunks, which introduces additional inference overhead compared to single-step policies. 
More efficient diffusion sampling strategies could reduce this overhead and make action-chunk generation more suitable for real-time scenarios.

PDPO currently follows a receding-horizon execution scheme by applying only the first action of each generated chunk. 
While this preserves responsiveness to dynamic crowds, future extensions could introduce trajectory-level objectives or consistency regularization to better supervise the entire action sequence.

So far, PDPO evaluation is performed in simulated scenarios, where human agents follow fixed policies and do not react to the robot. 
This enables controlled comparisons with prior crowd-navigation methods, but it remains a simulation-level correction and does not fully capture real human--robot interactions. 
Future work should evaluate PDPO in more diverse crowd scenarios, including real-world or higher-fidelity simulated environments.
%===============================================================================
\section{Conclusion}
\label{sec:conclusion}

In this work, we introduce Planning Diffusion Policy Optimization (PDPO) for robot crowd navigation. 
PDPO uses a diffusion policy to generate short-horizon action chunks and executes them in a receding-horizon manner, enabling the robot to represent diverse local motion choices while remaining responsive to dynamic crowd interactions.

Our results highlight the role of diffusion-based action chunking in dense crowd navigation. 
Across quantitative ablations and qualitative visualizations, PDPO benefits from representing navigation decisions as short-horizon action sequences rather than isolated single-step actions. 
The generated chunks form locally coherent motion patterns and produce diverse short-horizon proposals under the same observation, reflecting the multi-modal structure of crowd navigation.

In addition, our benchmark analysis shows that missing boundary constraints can affect policy evaluation, motivating the use of bounded environments for more faithful assessment of crowd-aware navigation behaviors. 
Overall, PDPO provides an effective diffusion-based short-horizon planning framework for dense crowd navigation and highlights the importance of careful benchmark design for learned navigation policies.
%===============================================================================

%===============================================================================

\clearpage

\bibliography{example}  % .bib

%===============================================================================
\clearpage
\appendix

\section{Diffusion Policy and PPO Fine-tuning Details}
\label{app:diffusion_ppo}

This section provides additional details on how the diffusion policy used in PDPO
is fine-tuned with Proximal Policy Optimization (PPO). We follow the general
view of diffusion policy optimization in which the denoising process is treated
as an internal Markov decision process and optimized jointly with the environment
MDP.

\subsection{Diffusion Policy as an Augmented MDP}
\label{app:diffusion_mdp}

Following diffusion-policy optimization~\citep{DBLP:conf/iclr/RenLAS0MBDS25},
we view the reverse denoising process as an internal decision process and
construct an augmented MDP over denoising transitions. This formulation allows
the diffusion policy to be optimized with policy-gradient methods while keeping
the environment-level MDP unchanged.

We consider the environment MDP introduced in Section~\ref{sec:problem_formulation},
\begin{equation}
    \mathcal{M}_{\mathrm{env}}
    =
    (\mathcal{S}, \mathcal{A}, \mathcal{P}_0, \mathcal{P}, \mathcal{R}),
\end{equation}
where \(s_t \in \mathcal{S}\) denotes the crowd-navigation state at environment
time \(t\), and \(a_t \in \mathcal{A}\) is the robot action executed in the
environment.

In PDPO, the policy does not output a single Gaussian action directly. Instead,
it generates an action chunk
\begin{equation}
    \mathbf{x}_0^t
    =
    \left(x_{0,1}^t, x_{0,2}^t, \ldots, x_{0,H}^t\right),
\end{equation}
where \(H\) is the action chunk length and each \(x_{0,j}^t \in \mathbb{R}^2\)
denotes the \(j\)-th velocity command in the chunk. In our experiments, we use
\(H=5\), corresponding to a short-horizon plan of \(1.25\,\mathrm{s}\). During
execution, we follow a receding-horizon control scheme: only the first action
\(a_t = x_{0,1}^t\) is applied to the environment, and a new action chunk is
generated at the next environment timestep.

The action chunk is generated through a \(K\)-step reverse diffusion process.
Starting from Gaussian noise,
\begin{equation}
    \mathbf{x}_K^t \sim \mathcal{N}(0, I),
\end{equation}
the policy iteratively denoises the action chunk according to
\begin{equation}
    p_\theta(\mathbf{x}_{k-1}^t \mid \mathbf{x}_k^t, s_t)
    =
    \mathcal{N}
    \left(
        \mathbf{x}_{k-1}^t ;
        \mu_\theta(\mathbf{x}_k^t, s_t, k),
        \sigma_k^2 I
    \right),
    \qquad k = K, \ldots, 1,
\end{equation}
where \(\mu_\theta\) is parameterized by the diffusion noise prediction network
\(\varepsilon_\theta\). The final denoised sample \(\mathbf{x}_0^t\) is used as
the policy output.

To optimize this policy with policy gradient methods, we construct an augmented
diffusion MDP
\begin{equation}
    \bar{\mathcal{M}}
    =
    (\bar{\mathcal{S}}, \bar{\mathcal{A}},
    \bar{\mathcal{P}}_0, \bar{\mathcal{P}}, \bar{\mathcal{R}}),
\end{equation}
where each augmented timestep corresponds to one denoising transition inside an
environment timestep. For environment time \(t\) and denoising index
\(k \in \{K, K-1, \ldots, 1\}\), we define the augmented state and action as
\begin{equation}
    \bar{s}_{t,k}
    =
    (s_t, \mathbf{x}_k^t, k),
    \qquad
    \bar{a}_{t,k}
    =
    \mathbf{x}_{k-1}^t.
\end{equation}
The corresponding augmented policy is
\begin{equation}
    \bar{\pi}_\theta(\bar{a}_{t,k} \mid \bar{s}_{t,k})
    =
    p_\theta(\mathbf{x}_{k-1}^t \mid \mathbf{x}_k^t, s_t).
\end{equation}

Within an environment timestep, transitions are given by the learned reverse
diffusion process. After the final denoising step \(k=1\), the first action in
the generated chunk, \(a_t = x_{0,1}^t\), is executed in the environment. The
environment then transitions according to
\begin{equation}
    s_{t+1} \sim \mathcal{P}(s_{t+1} \mid s_t, a_t),
    \qquad
    a_t = x_{0,1}^t,
\end{equation}
and the next denoising trajectory is initialized by sampling
\(\mathbf{x}_K^{t+1} \sim \mathcal{N}(0,I)\).

The environment reward is assigned only after the final denoising step:
\begin{equation}
    \bar{R}(\bar{s}_{t,k}, \bar{a}_{t,k})
    =
    \begin{cases}
        \mathcal{R}(s_t, a_t), & k = 1, \\
        0, & k > 1,
    \end{cases}
    \qquad
    a_t = x_{0,1}^t.
\end{equation}
This construction allows the multi-step denoising procedure to be interpreted as
a sequence of internal policy decisions whose final output determines the action
executed in the crowd-navigation environment.

\subsection{PPO Fine-tuning over Denoising Steps}
\label{app:ppo_denoising}

Following PPO~\citep{DBLP:journals/corr/SchulmanWDRK17} and its use for
diffusion-policy optimization~\citep{DBLP:conf/iclr/RenLAS0MBDS25}, we optimize
the augmented denoising policy at the level of individual denoising transitions.

Since the augmented process defines a valid stochastic policy over denoising
transitions, we can apply PPO to fine-tune the diffusion policy after behavioral
cloning pretraining. For each denoising transition, we store
\begin{equation}
    (\bar{s}_{t,k}, \bar{a}_{t,k},
    \log \bar{\pi}_{\theta_{\mathrm{old}}}(\bar{a}_{t,k} \mid \bar{s}_{t,k}),
    \hat{A}_{t,k}),
\end{equation}
where \(\theta_{\mathrm{old}}\) denotes the policy parameters used to collect the
rollout data, and \(\hat{A}_{t,k}\) is the advantage assigned to the denoising
step.

The PPO probability ratio is computed at the denoising-step level:
\begin{equation}
    r_{t,k}(\theta)
    =
    \frac{
    \bar{\pi}_{\theta}(\bar{a}_{t,k} \mid \bar{s}_{t,k})
    }{
    \bar{\pi}_{\theta_{\mathrm{old}}}(\bar{a}_{t,k} \mid \bar{s}_{t,k})
    }.
\end{equation}
The clipped PPO objective is then
\begin{equation}
    \mathcal{L}_{\mathrm{PPO}}(\theta)
    =
    \mathbb{E}_{t,k}
    \left[
    \min
    \left(
        r_{t,k}(\theta) \hat{A}_{t,k},
        \mathrm{clip}
        \left(
            r_{t,k}(\theta),
            1-\epsilon,
            1+\epsilon
        \right)
        \hat{A}_{t,k}
    \right)
    \right].
\end{equation}
In practice, we maximize this surrogate objective together with the standard
value-function and entropy terms:
\begin{equation}
    \mathcal{L}(\theta)
    =
    \mathcal{L}_{\mathrm{PPO}}(\theta)
    -
    c_v \mathcal{L}_{\mathrm{value}}
    +
    c_e \mathcal{H}(\bar{\pi}_\theta),
\end{equation}
where \(c_v\) and \(c_e\) are weighting coefficients.

\subsection{Advantage Assignment for Diffusion Policies}
\label{app:diffusion_advantage}

Following diffusion-policy optimization~\citep{DBLP:conf/iclr/RenLAS0MBDS25},
we assign environment-level advantages to the denoising transitions that produce
the executed action.

The environment reward is observed only after the denoised action is executed.
Therefore, the same environment-level return must be assigned to the denoising
transitions that produced the executed action. We first compute an
environment-level return
\begin{equation}
    G_t
    =
    \sum_{t' \ge t}
    \gamma_{\mathrm{env}}^{t'-t}
    \mathcal{R}(s_{t'}, a_{t'}),
\end{equation}
where \(a_{t'} = x_{0,1}^{t'}\) is the first action of the generated chunk at
environment time \(t'\), and \(\gamma_{\mathrm{env}}\) is the environment discount
factor. The value function is defined over the environment state only:
\begin{equation}
    V_\phi(s_t) \approx \mathbb{E}[G_t \mid s_t].
\end{equation}
This gives the environment-level advantage
\begin{equation}
    \hat{A}^{\mathrm{env}}_t
    =
    G_t - V_\phi(s_t).
\end{equation}

To distribute this advantage over the denoising steps, we use a denoising
discount factor \(\gamma_{\mathrm{denoise}}\). Earlier denoising steps operate
on noisier action chunks and are therefore assigned smaller weights, while later
denoising steps receive larger credit. Specifically, for denoising step
\(k \in \{K, \ldots, 1\}\), we define
\begin{equation}
    \hat{A}_{t,k}
    =
    \gamma_{\mathrm{denoise}}^{k-1}
    \hat{A}^{\mathrm{env}}_t.
\end{equation}
Thus, the final denoising transition \(k=1\) receives the full environment-level
advantage, while earlier transitions are downweighted according to their distance
from the final denoised action.

This advantage assignment provides a practical way to couple environment-level
reinforcement learning with the internal denoising process of the diffusion
policy. It allows PPO to update all denoising transitions involved in generating
an executed action, while giving higher weight to denoising steps that are closer
to the final action chunk.

\subsection{Behavioral Cloning Pretraining}
\label{app:bc_pretraining}

Following diffusion-policy imitation learning~\citep{DBLP:journals/ijrr/ChiXFCDBTS25}
and offline-to-online diffusion-policy optimization~\citep{DBLP:conf/iclr/RenLAS0MBDS25},
we pretrain the diffusion policy using behavioral cloning on demonstration
trajectories before online PPO fine-tuning.

Given an expert action chunk \(\mathbf{x}_0^t\), we sample a diffusion step \(k\)
and Gaussian noise \(\epsilon \sim \mathcal{N}(0,I)\), and construct the noisy
action chunk
\begin{equation}
    \mathbf{x}_k^t
    =
    \sqrt{\bar{\alpha}_k}\mathbf{x}_0^t
    +
    \sqrt{1-\bar{\alpha}_k}\epsilon.
\end{equation}
The denoising network is trained to predict the injected noise:
\begin{equation}
    \mathcal{L}_{\mathrm{BC}}(\theta)
    =
    \mathbb{E}_{t,k,\epsilon}
    \left[
        \left\|
        \epsilon -
        \epsilon_\theta(\mathbf{x}_k^t, s_t, k)
        \right\|_2^2
    \right].
\end{equation}
This pretraining stage provides a strong initialization that captures common
collision-avoidance behaviors before online reinforcement learning.

\section{Network Architecture and Hyperparameters}
\label{app:architecture_hyperparameters}

\subsection{Observation Encoder Architecture}
Our observation encoder follows prior crowd-navigation architectures,
including CrowdNav++~\citep{DBLP:conf/icra/LiuCHCHLMGD23} and
ASTG~\citep{DBLP:conf/icra/ZhouG24}, which use neural interaction modeling
to encode human–robot and human–human relationships.
The observation encoder maps the current robot state and the predicted human
states into a compact latent representation used to condition the diffusion
policy. At timestep \(t\), the input observation is
\[
s_t = [w^t, h_1^t, \dots, h_{n_t}^t],
\]
where \(w^t\) denotes the robot state and \(h_i^t\) denotes the feature of the
\(i\)-th observable human. Each human feature contains the current human position
and its constant-velocity predicted positions over the next \(L\) timesteps:
\[
h_i^t = [u_i^t, \hat{u}_i^{t+1:t+L}].
\]
In all experiments, we use \(L=5\). For batching, the number of humans is padded
to a fixed maximum number, and a visibility mask is used to ignore padded or
unobservable humans.

\paragraph{Human embedding and masked self-attention.}
Each human feature is first projected into a latent embedding using a multilayer
perceptron:
\[
h_i^{\mathrm{emb}} = \mathrm{MLP}_{\mathrm{human}}(h_i^t).
\]
To model interactions among nearby humans, we apply masked multi-head
self-attention over the set of human embeddings:
\[
\tilde{h}_1, \dots, \tilde{h}_{n_t}
=
\mathrm{SelfAttn}
\left(
h_1^{\mathrm{emb}}, \dots, h_{n_t}^{\mathrm{emb}}
\right).
\]
The attention mask prevents padded or unobservable humans from contributing to
the attention computation. Residual connections, layer normalization, and
dropout are used after the attention layer to stabilize optimization. The
resulting embeddings \(\tilde{h}_i\) encode interaction-aware human features.

\paragraph{Robot-conditioned human--robot features.}
To capture how each human relates to the robot, we concatenate the robot state
with each interaction-aware human embedding and process the result with an edge
MLP:
\[
e_i = \mathrm{MLP}_{\mathrm{edge}}([\tilde{h}_i, w^t]).
\]
This produces a set of robot-conditioned human--robot interaction features
\(\{e_i\}_{i=1}^{n_t}\).

\paragraph{Attention pooling over humans.}
Since the number of observable humans varies over time, we aggregate the
interaction features using attention pooling. For each human--robot feature
\(e_i\), an attention score is computed as
\[
q_i = \mathrm{MLP}_{\mathrm{att}}(e_i),
\]
and normalized across observable humans:
\[
\alpha_i =
\frac{\exp(q_i)}
{\sum_{j \in \mathcal{V}_t} \exp(q_j)},
\]
where \(\mathcal{V}_t\) denotes the set of visible humans at timestep \(t\). The
pooled crowd representation is then
\[
e^{\mathrm{att}} = \sum_{i \in \mathcal{V}_t} \alpha_i e_i.
\]
This attention-pooling mechanism allows the encoder to emphasize humans that are
most relevant to the robot's near-future motion.

\paragraph{Latent projection.}
Finally, the pooled crowd representation is concatenated with the robot state
and passed through a latent projection MLP:
\[
z_t = \mathrm{MLP}_{\mathrm{latent}}([w^t, e^{\mathrm{att}}]).
\]
The latent vector \(z_t\) is used as the conditioning input for both the
diffusion denoising network and the value function during online fine-tuning.

\subsection{Diffusion Policy and Value Function}

The diffusion policy operates on an action chunk
\[
% x_0^t = [a_t^1, a_t^2, \dots, a_t^H],
x_0^t = \left[x_{0,1}^t, x_{0,2}^t, x_{0,3}^t, \ldots, x_{0,H}^t\right],
\]
where each action \(x^t_{0,j} \in \mathbb{R}^2\) is a velocity command. The noisy
action chunk \(x_k^t\) is flattened and concatenated with the observation latent
vector \(z_t\) and a diffusion timestep embedding \(\psi(k)\). The timestep
embedding is produced using sinusoidal positional encoding followed by a small
MLP. The denoising network predicts the injected Gaussian noise:
\[
\hat{\epsilon} = \epsilon_\theta(x_k^t, z_t, k).
\]
The output dimension is \(2H\), matching the flattened action-chunk dimension.
For the single-action ablation, we use the same architecture with \(H=1\).

The value function shares the same observation encoder and predicts a scalar
value from the latent representation:
\[
V_\phi(s_t) = \mathrm{MLP}_{\mathrm{value}}(z_t).
\]
This value estimate is used to compute the environment-level advantage during
PPO fine-tuning, which is then assigned to the denoising transitions as described
in Appendix~\ref{app:diffusion_advantage}.

\subsection{Hyperparameters}

Table~\ref{tab:hyperparameters} summarizes the main hyperparameters used in our
experiments.

\begin{table}[!p]
\centering
\small
\caption{Environment settings, network architecture, diffusion-model settings, and training hyperparameters.}
\label{tab:hyperparameters}
\begin{tabular}{ll}
\toprule
\textbf{Hyperparameter} & \textbf{Value} \\
\midrule
Workspace size & $12 \times 12~\mathrm{m}^2$ \\
Number of humans & 20 \\
Robot radius & $0.2~\mathrm{m}$ \\
Robot maximum speed & $1.0~\mathrm{m/s}$ \\
Robot sensing range & $6.0~\mathrm{m}$ \\
Control interval & $0.25~\mathrm{s}$ \\
Episode time limit & $50~\mathrm{s}$ \\
Human prediction horizon $L$ & 5 \\
Observation dimension & 269 \\
Action dimension & 2 \\
Action chunk length $H$ & 5 \\
Policy output dimension & $2H = 10$ \\
Planning horizon & $1.25~\mathrm{s}$ \\
\midrule
Demonstrations & $3{,}000$ episodes \\
BC optimization iterations & 200 \\
BC batch size & 128 \\
BC learning rate & $1 \times 10^{-3}$ \\
BC weight decay & $1 \times 10^{-6}$ \\
Online fine-tuning budget & $12$M environment steps \\
Evaluation episodes & 500 unseen test seeds \\
\midrule
Denoising network MLP & $[512, 256, 256]$ \\
Diffusion timestep embedding dimension & 16 \\
Denoising activation & ReLU \\
Value network MLP & $[256, 256, 256]$ \\
Value activation & Mish \\
Diffusion objective & noise prediction \\
Noise schedule & cosine beta schedule \\
Diffusion pretraining steps & 20 \\
Diffusion fine-tuning steps & 10 \\
\midrule
BC optimizer & AdamW \\
PPO optimizer & AdamW \\
PPO actor learning rate & $1 \times 10^{-4}$ \\
PPO critic learning rate & $1 \times 10^{-3}$ \\
Discount factor $\gamma$ & 0.99 \\
GAE parameter $\lambda$ & 0.95 \\
Denoising discount factor $\gamma_{\mathrm{denoise}}$ & 0.99 \\
PPO batch size & 500 \\
PPO update epochs & 5 \\
Value loss coefficient & 0.5 \\
Denoising policy loss clipping coefficient & 0.01 \\
Base denoising policy loss clipping coefficient & 0.001 \\
Denoising policy loss clipping growth rate & 3 \\
Target KL & 1.0 \\
\bottomrule
\end{tabular}

\vspace{0.4em}
\footnotesize{
The denoising policy loss clipping coefficient is used in the clipped diffusion-policy loss during online fine-tuning and is distinct from the standard PPO probability-ratio clipping threshold.
}
\end{table}

\FloatBarrier

\subsection{Model Size}

We additionally report the number of trainable parameters for learning-based
methods. During online fine-tuning, PDPO updates \(1.049\)M trainable parameters,
including a \(0.848\)M-parameter fine-tuned diffusion actor and a \(0.201\)M-parameter critic.
CrowdNav++ contains \(2.503\)M trainable parameters under our implementation.
Thus, PDPO uses fewer trainable parameters than the learning-based baseline
while achieving higher success rate in \emph{Bounded CrowdNav}.

\begin{table}[!htbp]
\centering
\caption{Trainable parameter count of learning-based methods.}
\label{tab:model_size}
\begin{tabular}{lc}
\toprule
\textbf{Method} & \textbf{Trainable Parameters} \\
\midrule
CrowdNav++ (Const. Vel.) & $2.503$M \\
PDPO (Ours) & $1.049$M \\
\bottomrule
\end{tabular}
\end{table}

\FloatBarrier

\subsection{Training Curves}
\label{app:training_curves}

\begin{figure}[!htbp]
    \centering
    \includegraphics[width=0.47\linewidth]{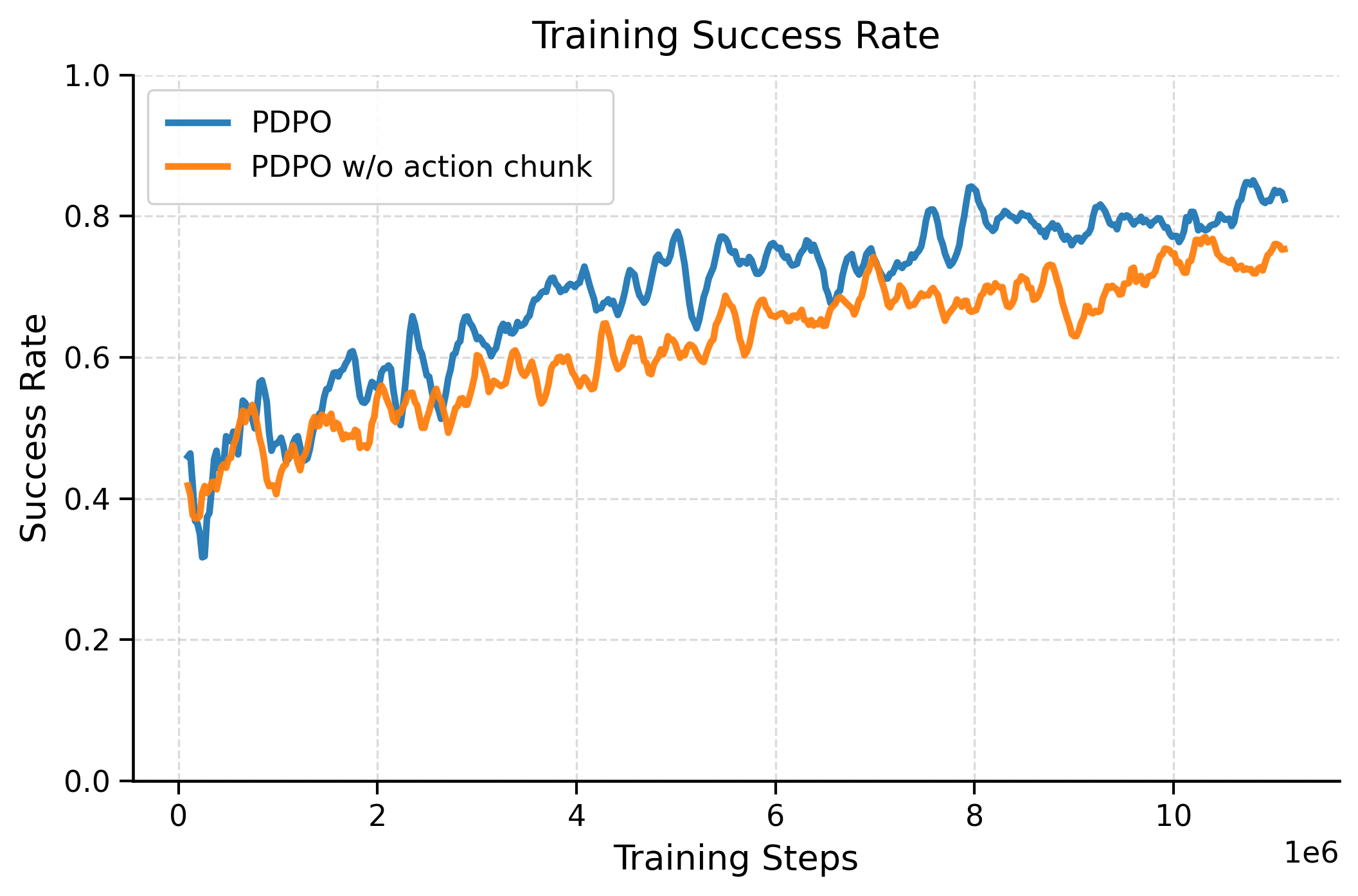}
    \hfill
    \includegraphics[width=0.47\linewidth]{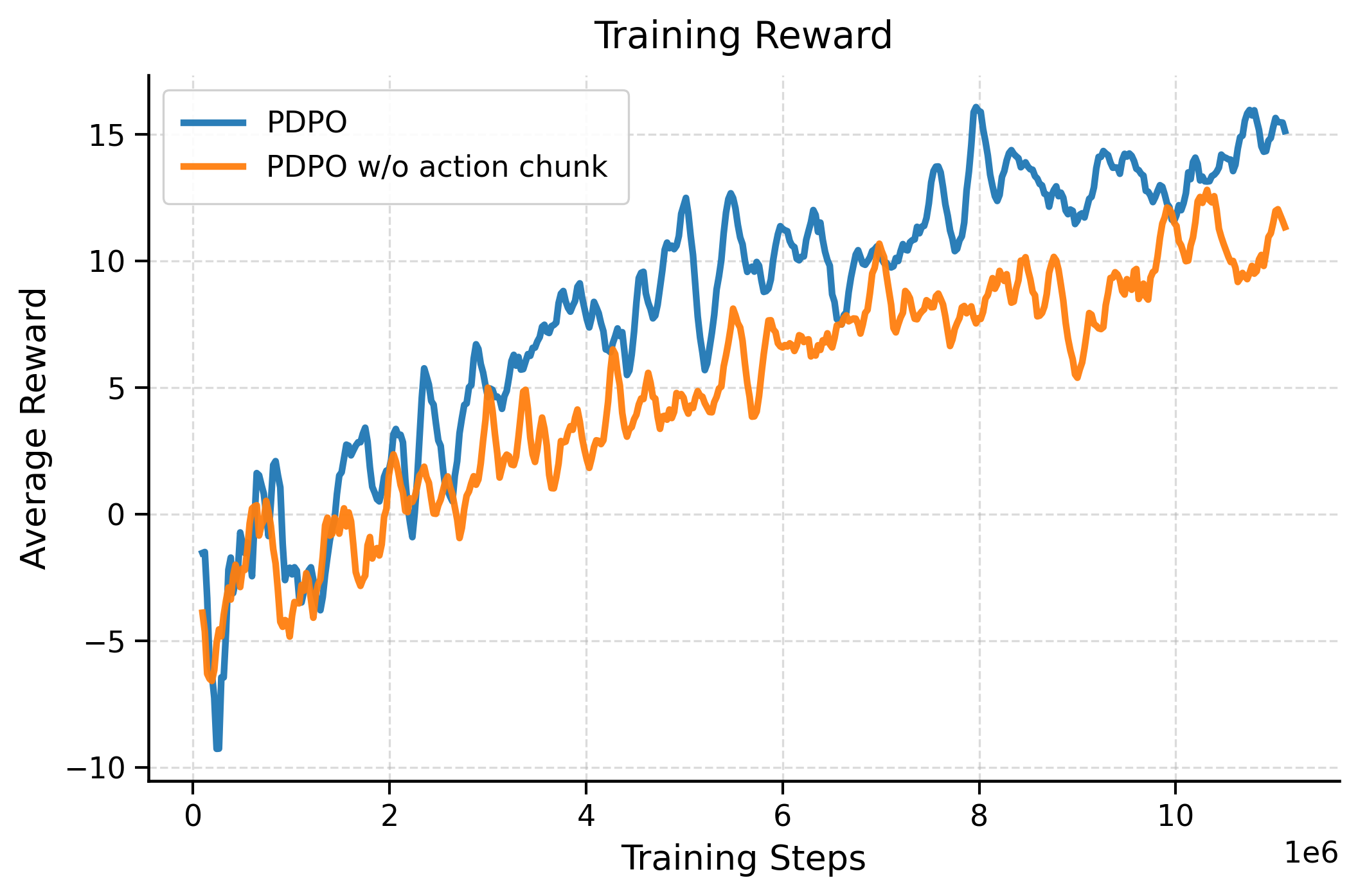}
    \caption{ Online fine-tuning curves of PDPO and its ablation on Bounded CrowdNav over the common training horizon. Left: training success rate. Right: average training reward. }
    \label{fig:training_curves_bounded}
\end{figure}

Figure~\ref{fig:training_curves_bounded} shows the online fine-tuning curves on
Bounded CrowdNav. Compared with the single-action diffusion-policy ablation,
PDPO achieves higher and more stable training success rates, supporting the
benefit of action-chunk generation during online fine-tuning. For the final test evaluation, we select the checkpoint that achieves the
highest success rate on the validation episodes during training.

\FloatBarrier

\section{Additional Visualization of Action Chunks}
\label{app:action_chunk_visualization}

This section provides additional qualitative visualizations of action chunks
generated by PDPO in dense crowd-navigation scenarios. Each snapshot is centered
on the robot: the dashed circle denotes the observation range, blue dots denote
humans with predicted linear trajectories, the green dot denotes the robot, and
the red cross denotes the end of the action chunk. At each decision step, PDPO
generates a 5-step action chunk with a control interval of \(0.25\,\mathrm{s}\),
corresponding to a \(1.25\,\mathrm{s}\) short-horizon plan. Only the first action
is executed before replanning at the next timestep.

\begin{figure}[!p]
    \centering
    \begin{tabular}{@{}ccc@{}}
        \includegraphics[width=0.3\textwidth]{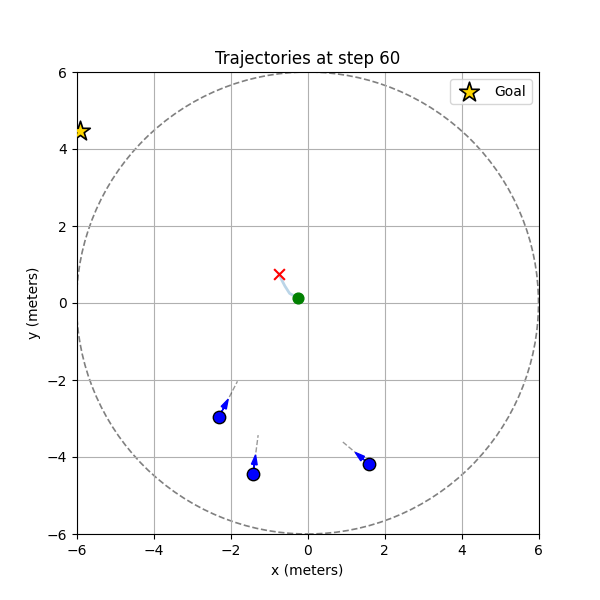} &
        \includegraphics[width=0.3\textwidth]{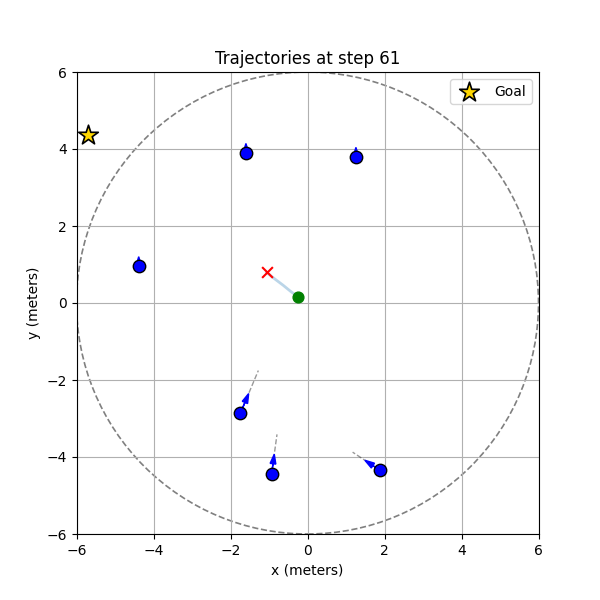} &
        \includegraphics[width=0.3\textwidth]{figures_of_action_chunks_examples/step_0062.png} \\
        \small Step 60 & \small Step 61 & \small Step 62 \\[0.4em]
        \includegraphics[width=0.3\textwidth]{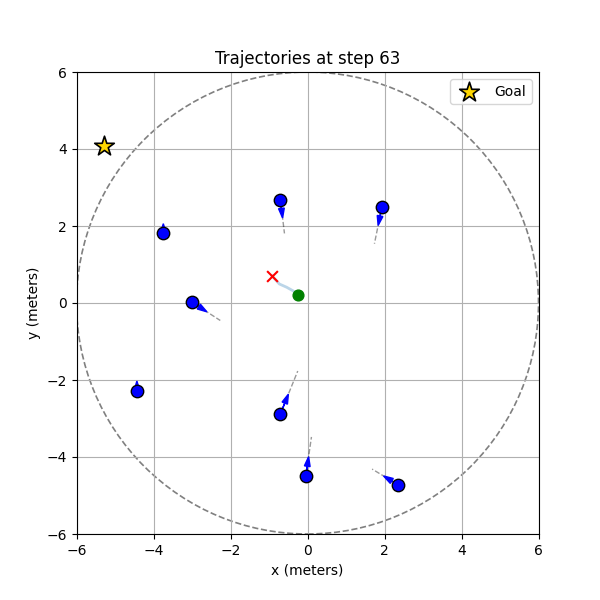} &
        \includegraphics[width=0.3\textwidth]{figures_of_action_chunks_examples/step_0064.png} &
        \includegraphics[width=0.3\textwidth]{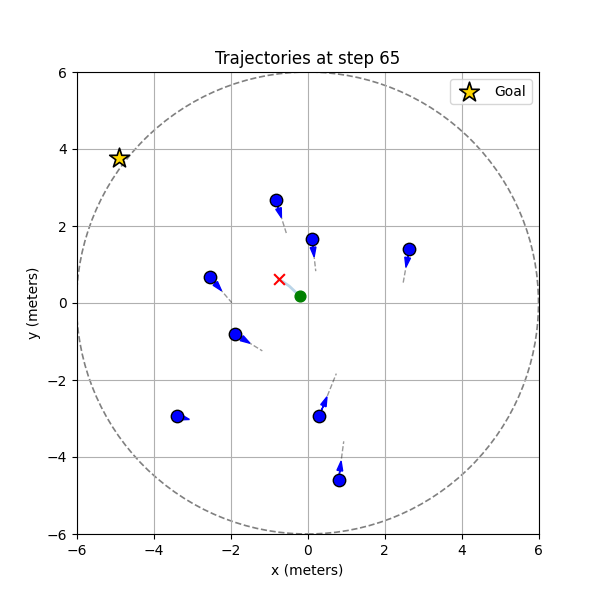} \\
        \small Step 63 & \small Step 64 & \small Step 65 \\[0.4em]
        \includegraphics[width=0.3\textwidth]{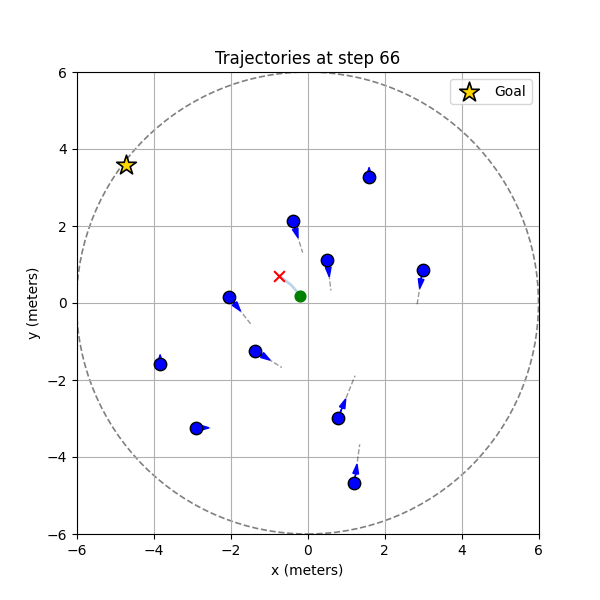} &
        \includegraphics[width=0.3\textwidth]{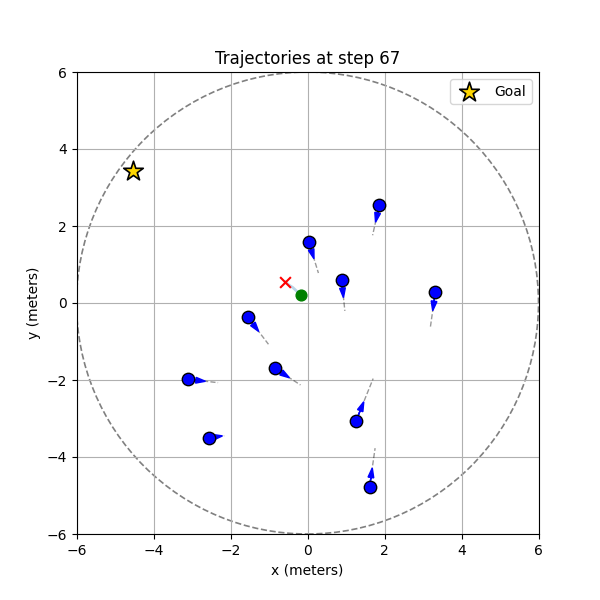} &
        \includegraphics[width=0.3\textwidth]{figures_of_action_chunks_examples/step_0068.png} \\
        \small Step 66 & \small Step 67 & \small Step 68 \\
    \end{tabular}
    \caption{
    Complete storyboard of a PDPO trajectory from step 60 to step 68. The
    generated action chunks remain coherent while adapting to nearby pedestrians.
    }
    \label{fig:app_action_chunks_coherence}
\end{figure}

Figure~\ref{fig:app_action_chunks_coherence} shows an episode where the robot
moves through a dense crowd toward a northwest goal. As pedestrians enter the
observation range, the generated chunks adjust toward available gaps while
maintaining clearance. Around steps 62--64, the plan changes smoothly to turn
around approaching pedestrians. Steps 65--67 show shorter segments in the dense
region, indicating more cautious motion; after the robot leaves this constrained
area, the planned motion becomes longer again.

\begin{figure}[!p]
    \centering
    \begin{tabular}{@{}ccc@{}}
        \includegraphics[width=0.3\textwidth]{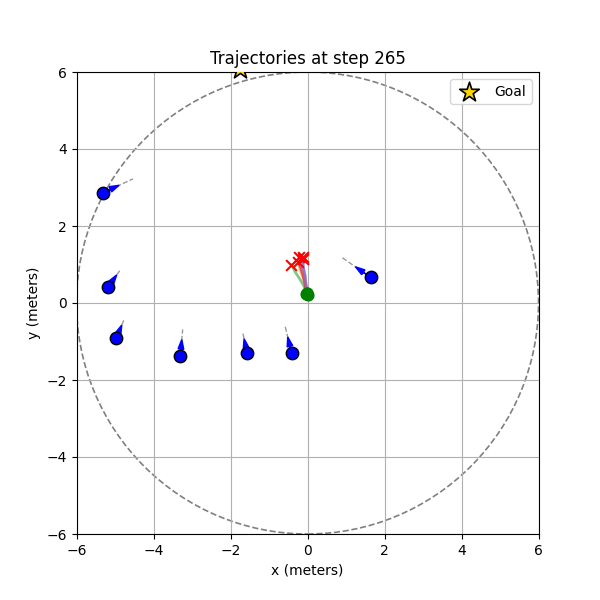} &
        \includegraphics[width=0.3\textwidth]{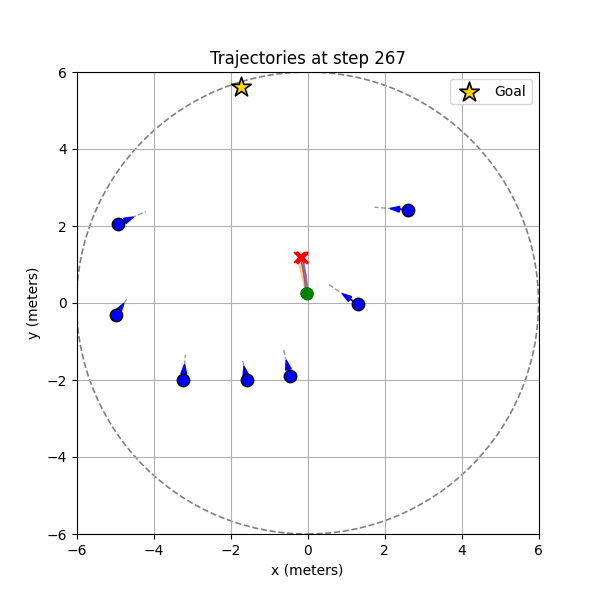} &
        \includegraphics[width=0.3\textwidth]{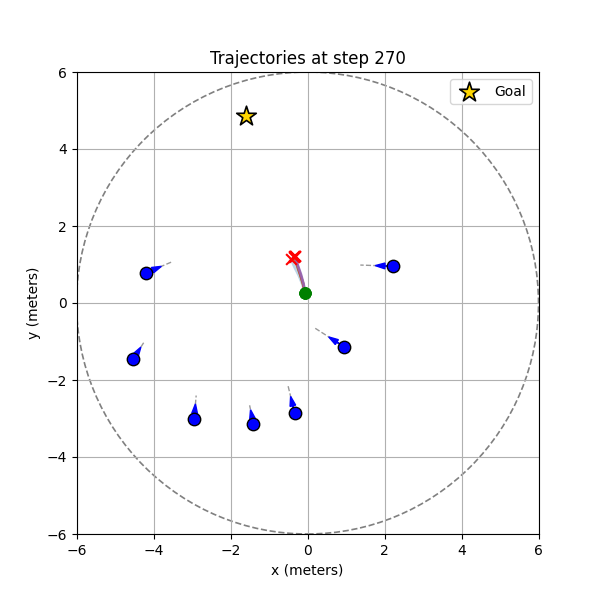} \\
        \small Step 265 & \small Step 267 & \small Step 270 \\[0.4em]
        \includegraphics[width=0.3\textwidth]{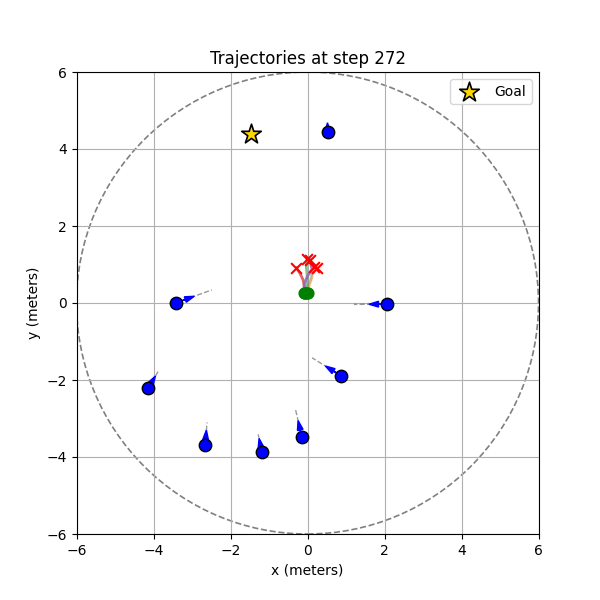} &
        \includegraphics[width=0.3\textwidth]{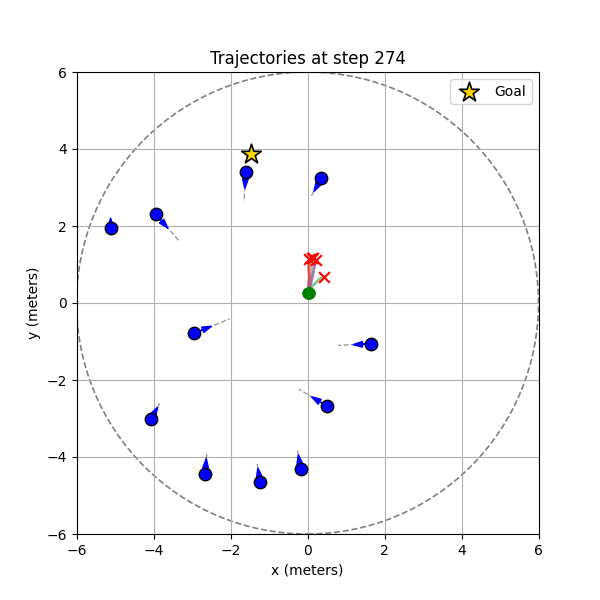} &
        \includegraphics[width=0.3\textwidth]{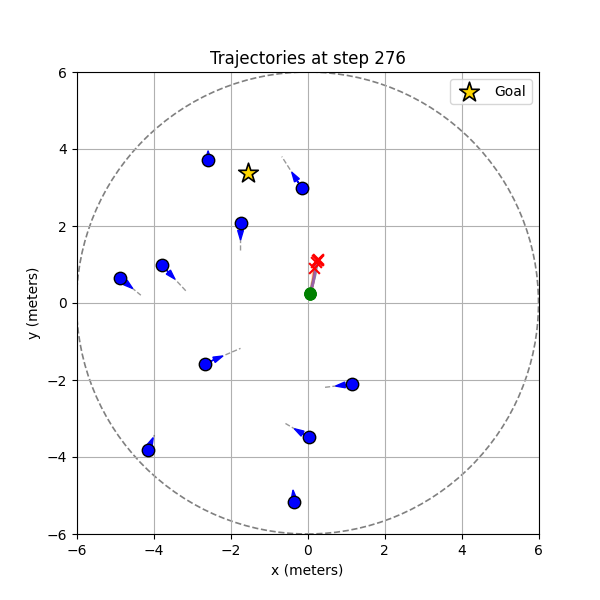} \\
        \small Step 272 & \small Step 274 & \small Step 276 \\[0.4em]
        \includegraphics[width=0.3\textwidth]{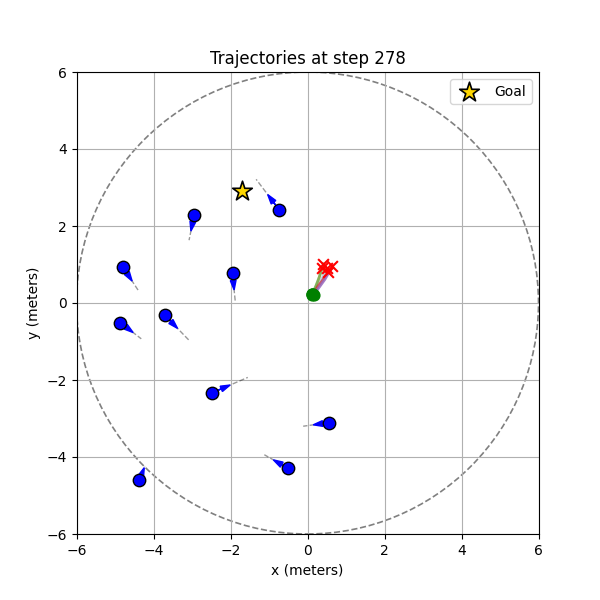} &
        \includegraphics[width=0.3\textwidth]{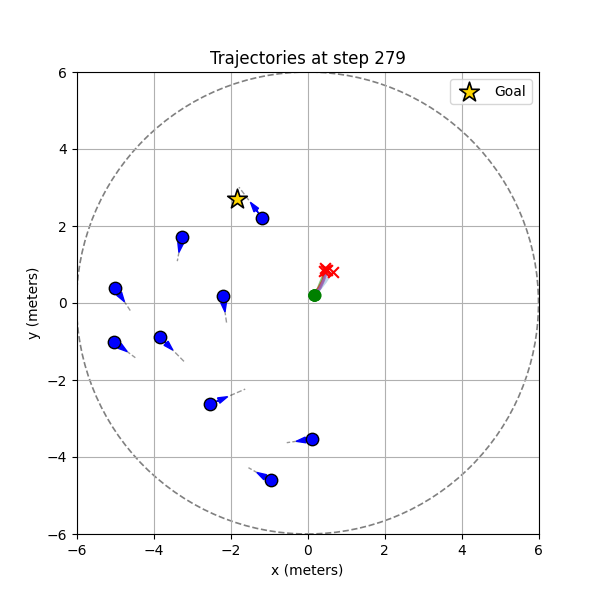} &
        \includegraphics[width=0.3\textwidth]{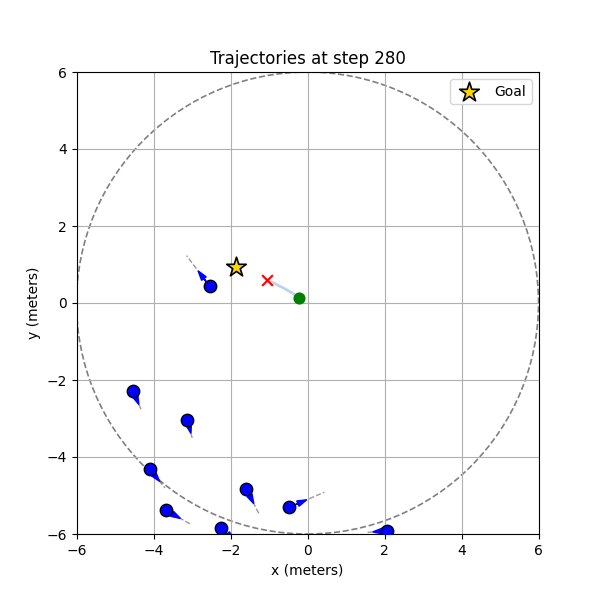} \\
        \small Step 278 & \small Step 279 & \small Step 280 \\
    \end{tabular}
    \caption{
    Complete storyboard of multimodal action chunks generated by PDPO. Multiple
    sampled short-horizon trajectories are visualized under the same observation,
    illustrating diverse feasible local plans.
    }
    \label{fig:app_action_chunks_multimodal}
\end{figure}

Figure~\ref{fig:app_action_chunks_multimodal} shows another dense-crowd episode.
At each timestep, multiple action chunks are sampled from the learned diffusion
policy under the same observation. These samples correspond to different feasible
short-horizon choices, such as passing through different gaps or slowing down
near pedestrians. This variability shows that PDPO represents multiple plausible
local plans and replans as the crowd evolves.

\FloatBarrier

\end{document}